# Production System Rules as Protein Complexes from Genetic Regulatory Networks

Larry Bull

Department of Computer Science & Creative Technologies,

University of the West of England, Bristol BS16 1QY, U.K.

Larry.Bull@uwe.ac.uk

**Abstract**

This short paper introduces a new way by which to design production system rules. An indirect encoding scheme is presented which views such rules as protein complexes produced by the temporal behaviour of an artificial genetic regulatory network. This initial study begins by using a simple Boolean regulatory network to produce traditional ternary-encoded rules before moving to a fuzzy variant to produce real-valued rules. Competitive performance is shown with related genetic regulatory networks and rule-based systems on benchmark problems.

# 1. Introduction

The genetic regulatory networks (GRN) within cells synthesize proteins, some of which have regulatory functions, some have intra-cellular function, and some pass through the cell membrane. A growing body of work incorporates increasing levels of detail from the natural phenomena for computational intelligence but very few have considered proteins explicitly, and only one is known to do so whilst also exploiting an underlying GRN [Knibbe et al., 2008]. In this paper, a well-known Boolean GRN model [Kauffman, 1969] is extended to consider the role of non-regulatory proteins in a simple way, in particular their formation of complexes. Such protein aggregations are viewed as production system rules and hence a new, indirect encoding for Learning Classifier Systems (LCS) [Holland, 1976] is presented. Whilst a number of indirect encodings have been presented for artificial neural networks (e.g., see [Yao, 1999][Floreano et al. 2008] for overviews) no prior work for rules is known. Initial results indicate that increases in performance are possible through the extra layer of abstraction in comparison to the traditional GRN model but not to the related Pittsburgh-style LCS [Smith, 1980]. The Boolean logic GRN model is then altered to one using simple fuzzy logic operations [Kok & Wang, 2006] to enable the design of rules for continuous-valued problems. Again using versions of well-known benchmark problems, it is shown that improved performance is obtained in comparison to equivalent Pittsburgh-style LCS.

It has recently been shown [Bull, 2012a] that a population-based algorithm which uses imitation as inspiration for its search mechanisms, as opposed to genetic evolution, is highly effective in the design of (dynamical) networks and so is used here.

# 2. Background

**2.1 Random Boolean Networks**

Within the traditional form of Random Boolean Networks (*RBN*) [Kauffman, 1969] there is a network of $R$ nodes, each with $B$ directed connections randomly assigned from other nodes in the network. All nodes update synchronously based upon the current state of those $B$ nodes. Hence those $B$ nodes are seen to have a regulatory effect upon the given node, specified by the given Boolean function randomly attributed to it. Nodes can also be self-connected. Since they have a finite number of possible states and they are deterministic, such networks eventually fall into an attractor. It is well-established that the value of $B$ affects the emergent behaviour of *RBN* wherein attractors typically contain an increasing number of states with increasing $B$. Three phases of behaviour were originally suggested through observation: ordered when $B=1$, with attractors consisting of one or a few states; chaotic when $B>3$, with a very large number of states per attractor; and, a critical regime around $1<B<4$, where similar states lie on trajectories that tend to neither diverge nor converge (see [Kauffman, 1993] for discussions of this critical regime, e.g., with respect to perturbations). Subsequent formal analysis using an annealed approximation of behaviour identified $B=2$ as the critical value of connectivity for behaviour change [Derrida & Pomeau, 1986]. Figure 1 shows examples of typical behaviour.

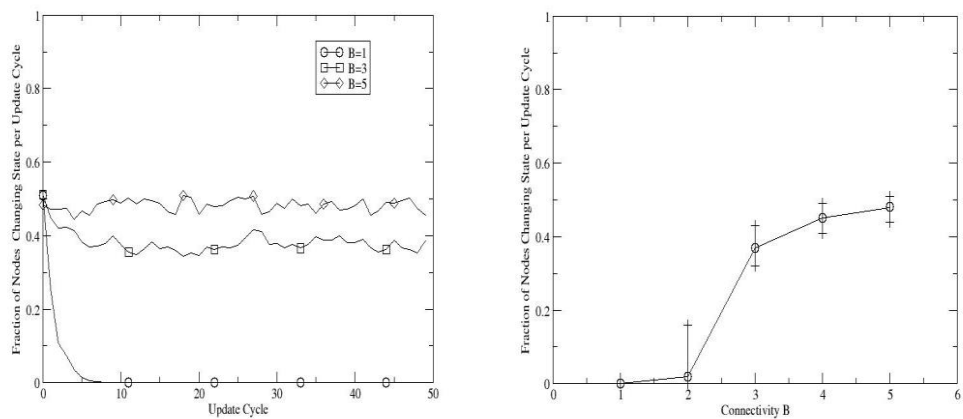

**Figure 1:** Typical behaviour of a traditional *RBN* with $R=1500$ nodes: on the left, showing example temporal dynamics; and on the right, the average behaviour (100 runs) after 100 update cycles. Nodes were initialized at random. Error bars show max and min behaviour.

There is a small amount of prior work exploring the use of computational intelligence to design *RBN*. Van den Broeck and Kawai [1990] used a simulated annealing-type approach to design feedforward *RBN* for the four-bit parity problem. Kauffman [1993, p.211] evolved *RBN* to match a given attractor (see also Lemke et al. [2001]). The same approach has been used to explore attractor stability [Fretter et al., 2009] and to model real regulatory network data, e.g., see [Tan & Tay, 2006]. Sipper and Ruppin [1997] evolved *RBN* for the well-known density task. Most closely related is work on the use of *RBN* to represent the rules within modern forms of Michigan-style LCS (e.g., [Preen & Bull, 2009]). In contrast to all this work, the approach presented here does not use the *RBN* to provide outputs from given inputs directly.

## 2.2 Protein-inspired Computation

The idea to consider cytoplasm proteins as a significant component in cellular information processing is not new [Paton, 1993][Bray, 1995] but little work in the area of computational intelligence appears to exist. This is, of course, not the same as the work on protein structure prediction, e.g., using computational intelligence, or the more specific modelling work of many aspects of cells typically undertaken in Systems Biology. Fisher et al. [1999] viewed proteins in signalling pathways as agents capable of rudimentary pattern recognition, memory, signal integration, etc. More recently, Qadir et al. [2010] have used a protein metaphor to realize relatively fault-tolerant associative memory in evolvable hardware. Most similar to the work presented here, and the only other known abstract model of genome and proteome interaction which may potentially be used as a general representation, is that by Knibbe et al. [2008]. They used a fuzzy representation wherein proteins of overlapping membership functions interact: proteins are produced by an underlying Gray encoded GRN and contribute fractionally to phenotypic traits based upon the degree of overlap. The model was used to explore effects of protein interactions on emerging genome structures. This paper views proteins as rule components.

## 2.4 Imitation Programming: Culture-inspired Search

The basic principle of imitation computation is that individuals alter themselves based upon another individual(s), typically with some error in the process. Individuals are not replaced with the descendants of other individuals as in evolutionary search; individuals persist through time, altering their solutions via imitation. Thus imitation may be cast as a directed stochastic search process, thereby combining aspects of both recombination and mutation used in evolutionary computation. Imitation Programming (IP) [Bull, 2012a] is such a population-based stochastic search process shown to be competitive with related evolutionary search:

```
BEGIN
INITIALISE population with random candidate solutions
EVALUATE each candidate
REPEAT UNTIL (TERMINATION CONDITION) DO
        FOR each candidate solution DO
                SELECT candidate(s) to imitate
                CHOOSE component(s) to imitate
                COPY the chosen component(s) with ERROR
                EVALUATE new solution
                REPLACE IF (UPDATE CONDITION) candidate with new solution
        OD
OD
END
```

In this paper, similar to Differential Evolution [Storn & Price, 1997], each individual in the population *P* creates one variant of itself and it is adopted if better per iteration. Other schemes are, of course, possible, e.g., Particle Swarm Optimization (PSO) [Kennedy & Eberhart, 1995] always accepts new solutions but then also "imitates" from the given individual's best ever solution per learning cycle. This aspect of the approach, like many others, is open to future investigation. The individual to imitate is chosen using a roulette-wheel scheme based on proportional solution utility, i.e., the traditional reproduction selection scheme used in Genetic Algorithms (GA) [Holland, 1975]. Again, other schemes, such as the spatial networks of PSO, could be used. Further details of how IP is used to design *RBN* are given below.

## 3. Rules from GRN

The *RBN* model is an abstraction of gene regulation and thus does not explicitly consider the role of proteins which are maintained within the cell body. It is those proteins which determine the primary response of the cell to its environment. Typically, such proteins also form multi-protein complexes of two or more proteins which can, for example, increase their catalytic capabilities. Of course, the environment has causal effects on gene expression through a series of protein-protein interactions but the timescale is significantly longer, typically around five minutes. That is, cells often respond to a given stimulus through protein complexes which were created by their GRN *before* the event. As noted above, in all the aforementioned previous work using *RBN* for computation and most artificial GRN research (e.g., [Bull, 2012b]), it is the slower response mechanism that is considered. That is, much like a form of neural network, problem inputs are encoded and applied to the GRN before it is updated to determine a response. In this paper, the formation of protein complexes and their use to provide responses to problem inputs has been considered through a simple extension, essentially resulting in an indirect encoding scheme for Pittsburgh-style LCS.

Within *RBN*, the expression of genes, and hence implicitly the formation of the protein they each encode, is considered a binary event. Thus if a gene node within a network has a logical value '1' it is considered 'on' and to have therefore produced a protein, and vice versa. To add a layer of protein complexes in a relatively simple but computationally useful form, in keeping with the *RBN* model, ternary-encoded production system rules are initially used. Here, for a set of predefined nodes in an *RBN*, the presence or absence of the protein expressed by a given gene is used to specify one part of a rule/complex. The series of states of the set of *RBN* nodes can therefore be used to specify multiple rules - for *T* update cycles, *T* rules can be determined.

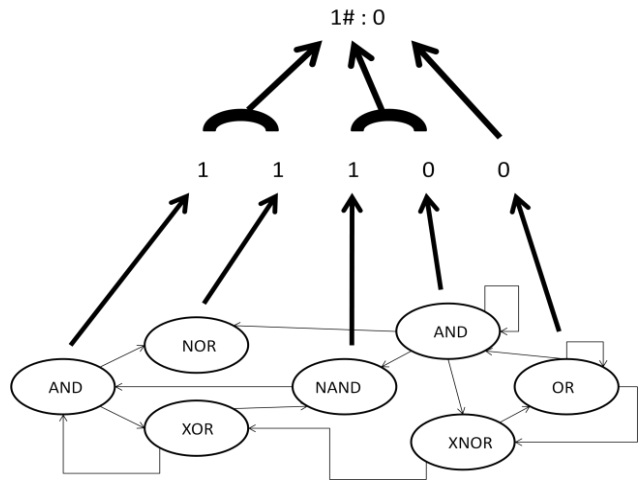

**Figure 2:** Showing how a production system rule is determined from an *RBN* per update cycle of the network. The current state of a set of predetermined genes is found and combined, first into a binary string and then into a ternary rule

Each gene node of a given network is set to its defined start state. Nodes then update synchronously for *T* cycles. On each cycle, the value of a set of *L* identified nodes is joined into a binary string which is subsequently turned into a traditional ternary LCS rule. Thus if the given problem has two binary-encoded inputs *I* and one binary output *O*, then *L*=5. The value of each of the first 2\**I* nodes are used to specify the rule condition. Each pair of bits is interpreted as either '0', '1' or the generalization symbol '#': '11' = '1', '00'='0', otherwise '#'. Actions are interpreted directly. Figure 2 shows an example. Once the *RBN* has been iterated for *T* cycles and the *T* rules determined, the fitness of the *RBN* is ascertained by evaluating the rules on the given problem. To avoid the issue of multiple rules matching a given input, the rules are ordered based upon the *RBN* cycle that created them. For example, if the rule created on *RBN* cycle 2 and that on cycle 8 both match a given input the action of the rule from cycle 2 will be used as the output. Of course, other schemes are possible but not considered here.

## 4. Experimentation I: Boolean Logic

**4.1 Imitation Programming *RBN***

For *RBN* design, IP utilizes a variable-length representation of pairs of integers defining node inputs, each with an accompanying single bit defining the node's start state, and an integer to define the node function, and there is an integer per *RBN* to define *T*. Five imitation operators are used: copy a node connection, copy a node start state, copy a node function, copy a network cycle count *T*, and change size through copying. In this paper, each operator can occur with or without error, with equal probability, such that an individual performs one of the ten during the imitation process as follows:

To copy a node connection, a randomly chosen node has one of its randomly chosen connections set to the same value as the corresponding node and its same connection in the individual it is imitating. When an error occurs, the connection is set to the next or previous node in the individual being imitated (equal probability, bounded by solution size). Imitation can also copy the start state for a randomly chosen node from the corresponding node, or do it with error (bit flip here). Size is altered by adding or deleting nodes and depends upon whether the two individuals are the same size. If the individual being imitated is larger than the copier, the connections and node start state of the first extra node are copied to the imitator, a randomly chosen node being connected to it. If the individual being imitated is smaller than the copied, the last added node is cut from the imitator and any/all connections to it re-assigned at random. If the two individuals are the same size, either event can occur (with equal probability). Node addition adds a randomly chosen node from the individual being imitated onto the end of the copier and it is randomly connected into the network. The operation can also occur with errors such that copied connections are either incremented or decremented. Deletion is as before. For a problem with a given number of inputs *I* and outputs *O*, the node deletion operator has no effect if the parent consists of $O+(2*I) = L$ nodes. Similarly, there is a maximum size (100) defined beyond which the growth operator has no effect. The number of cycles *T* can also be imitated, its value being incremented or decremented by 1 in the allowed range.

## 4.2 Results

The well-known benchmark multiplexer task is used in this paper since they can be used to build many other logic circuits, including larger multiplexers. These Boolean functions are defined for binary strings of length $l = x + 2^x$ under which the $x$ bits index into the remaining $2^x$ bits, returning the value of the indexed bit. The inverse task, i.e., the demultiplexer, is also used here. Upon each evaluation, each node in an *RBN* has its state set to its specified start state. The *RBN* is then executed for $T$ cycles, as encoded alongside the *RBN* topology. The value on the $L$ output nodes is recorded on each cycle and the production system rule formed and placed in an ordered list as described above. Using the rules produced by the *RBN*, the correct response to an input results in a quality increment of 1, with all possible $2^l$ binary inputs being presented per evaluation. All results presented are the average of 20 runs, with a population/society of μ=50 and $B$=2 (see Section 2.1). Nodes functions are from the allowed set {AND, NAND, OR, NOR, XOR, XNOR} and the number of rules allowed is 1≤$T$≤32, initialized uniform randomly. The *RBN* contain $L$ nodes initially. Adoption/replacement is based on quality, or $R$ if that is equal, or $T$ if $R$ is also equal, the smaller in both cases, or finally the decision is random if all three are equal.

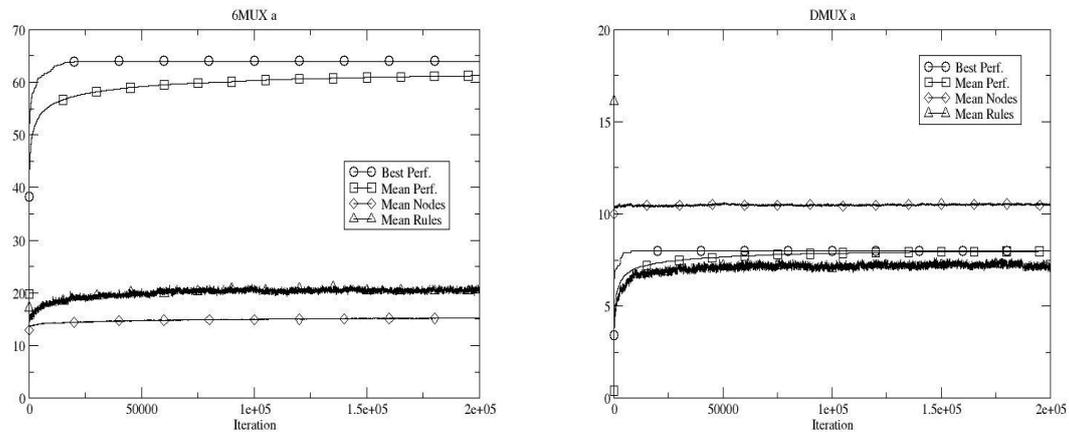

**Figure 3.** Performance on multiplexer (left) and demultiplexer (right).

Figure 3 (left) shows the performance of the approach on the 6-bit ($x$=2) multiplexer problem. Optimal performance (64) is obtained around 20,000 iterations, with 20 rules per individual. Figure 3 (right) shows performance of the same algorithm for an $x$=2 demultiplexer, i.e., one with three inputs and four outputs. It can be seen that optimal performance (8) is reached around 8,000 iterations, with 7-8 rules.

As noted in Section 2.1, *RBN* are discrete dynamical systems with a finite number of possible states and they are deterministic, hence such networks eventually fall into a basin of attraction. As Figure 1 (left) shows, for low *B* it typically takes around 10-15 update cycles for an attractor to be reached. This is potentially significant for the scheme described so far since it implies there is a finite set of different rules an *RBN* can produce, i.e., those from the states encountered into an attractor and then those produced (repeatedly) within it, regardless of the value of *T*. For example, for *B*=2, as used here, the typical length of an attractor is $\sqrt{R}$ [Kauffman, 1993] – a potentially small number for the *RBN* evolved here.

It should be noted that it is not an issue for the two tasks used above since they can be solved optimally with fewer than 10-15 rules. To reduce this general limitation, the *RBN* can be supplied with a changing external input to (potentially) keep it out of attractors. Figure 4 shows results for the tasks above but with the first five nodes of an *RBN* having their first connection receive the corresponding bit of a binary-encoded "clock" input. That is, since $1 \leq T \leq 32$, on each update cycle of the *RBN*, the corresponding binary pattern from the $2^5$ possible inputs is applied. As can be seen, in both cases there is no significant difference (T-test, p>0.05) in time taken to reach the optimum or in the size of the *RBN*, but there is a significant (T-test, p≤0.05) increase in the number of rules produced on the multiplexer (not the demux). This is perhaps a somewhat expected potential side-effect of keeping the *RBN* out of attractors.

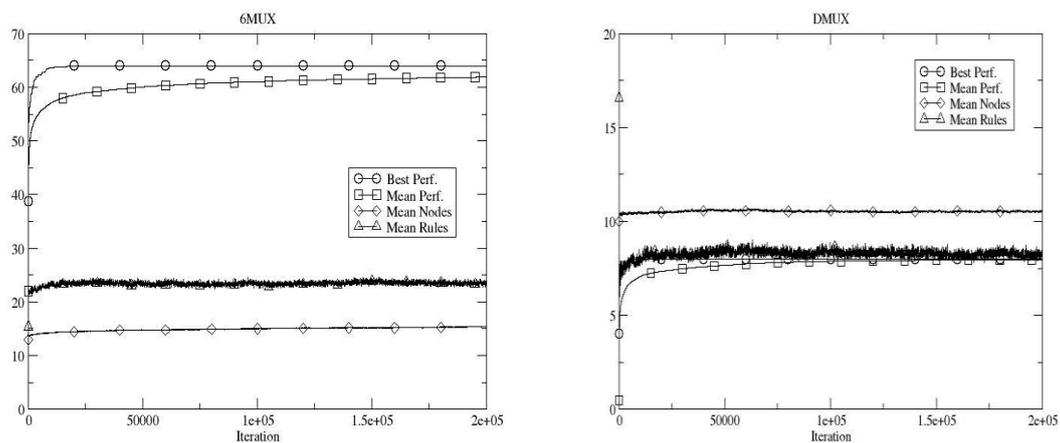

**Figure 4.** Performance on multiplexer (left) and demultiplexer (right) with a clock input applied.

As noted above, the standard use of *RBN* considers them as a form of recurrent neural network with inputs applied at some nodes, as is the clock input in the approach here, and outputs taken from other nodes, as the rule components are here. The performance of this approach has been explored using the same IP mechanisms as before (Section 4.1). Figure 5 shows how it is significantly slower (T-test, p≤0.05) to solve the multiplexer than the protein-based scheme (Figure 4). The same was true for the demuliplexer (not shown). And this is the case whether initialization gives the networks the minimum required $O+I$ nodes for the approach or $O+2I$ as above (the former is shown).

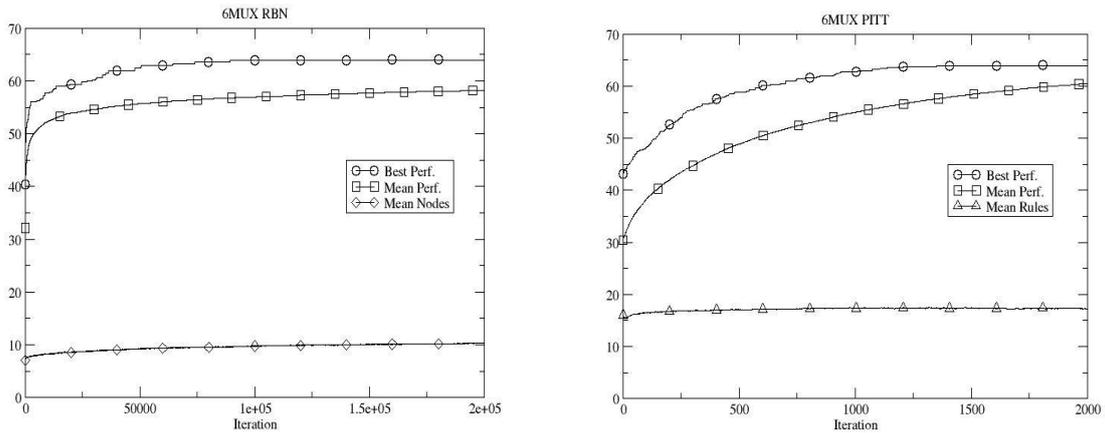

**Figure 5.** Performance on multiplexer of standard *RBN* approach (left) and direct encoding (right).

Figure 5 also shows the performance of using IP to design the standard, directly encoded rule-concatenation, ternary representation of Pittsburgh-style LCS. Here imitation can copy a rule component, or copy/delete a whole rule causing a change in size using the same general scheme described in Section 4.1. Each can occur with error and one of the four possible operations is applied with equal probability per iteration. Each individual was seeded with a number of rules uniform randomly in the same range as *T* above. As can be see, the standard representation is approximately ten times faster at finding the optimum on the multiplexer than the protein approach (Figure 4) using the same number of rules, and a similar result was found for the demultiplexer (not shown). However, this is perhaps not very surprising, particularly given the complexity and size of the *RBN* representation compared to the standard representation here: an *RBN* is defined by ω nodes, each encoded as 4 integers of various ranges, plus an extra integer for *T*; and, a directly encoded ruleset consists of ψ rules, each encoded as *I* ternary numbers and *O* binary bits. Note also that *(2I+O)*≤ω≤100 and 1≤ ψ≤100.

The following section shows how the indirect encoding can be beneficial when real-valued tasks are considered.

# 5. Experimentation II: Fuzzy Logic

## 5.1 Fuzzy Logic Networks

The continuous-valued dynamical systems known as Fuzzy Logic Networks (*FLN*) [Kok & Wang, 2006] are an extension of *RBN* where the Boolean functions are replaced with simple fuzzy logic functions. Kok and Wang explored 3-gene regulation networks using *FLN* and found that not only were *FLN* able to represent the varying degrees of gene expression but also that the dynamics of the networks were able to mimic a cell's irreversible changes into an invariant state or progress through a periodic cycle. A number of different fuzzy logic sets have been introduced since the original Max/Min method was proposed. Table 1 shows the six functions used here, again with $B=2$. Most closely related is work on the use of *FLN* to represent the rules within modern forms of Michigan-style LCS [Preen & Bull, 2011] also using the same function set. All other aspects remain the same as above, except start states are seeded uniform randomly [0.0,1.0] and their error is from the range [-0.1, 0.1] under imitation. The binary clock input is used.

**Table 1.** Fuzzy logic functions used by each node in the *FLN*.

| ID | Function | Logic |
|---|---|---|
| 0 | Fuzzy AND (Max/Min) | $\max(x,y)$ |
| 1 | Fuzzy AND (CFMQVS and Probabilistic) | $x*y$ |
| 2 | Fuzzy OR (Max/Min) | $\min(x,y)$ |
| 3 | Fuzzy OR (CFMQVS and MV) | $\min(1,x+y)$ |
| 4 | Fuzzy NOT | $1-x$ |
| 5 | Identity | $x$ |

Hence *FLN* enable the protein approach to produce real-valued rules from a *mostly integer-based* encoding. The GRN again requires $O+2I$ nodes as a minimum, where the two real-valued numbers per problem input variable represent the upper and lower bound of an (ordered) interval, e.g., see [Stone & Bull, 2003].

## 5.2 Results

The "real multiplexer" problem [Wilson, 2000] is used here which is an extension of the Boolean multiplexer: the binary input strings are replaced as real-valued vectors in the range [0,1]. Each value in the vector is then interpreted as 0 if greater than a threshold value, $\theta$, else 1, where typically $\theta=0.5$ as here, before being treated in the same way as the Boolean case for evaluation. Since there is no longer a finite set of possible inputs, a training set of 1000 randomly created vectors of length $l$ was created per experiment, together with another 1000 randomly created examples to act as the test set. The same approach was used to create a real-valued demultiplexer problem. All other parameters/details remained the same above. Figure 6 (left) shows the performance of the approach on the $x$=2 real multiplexer problem. Over the allowed time, performance is around 95% on the training set and 94% on the unseen test set, with around 13 rules and 17 nodes per individual, which is competitive with other reported results [Wilson, 2000]. Figure 6 (right) shows performance of the same algorithm for an $x$=2 real demultiplexer. It can be seen that performance is around 97% on the training set and 87% on the test set, with typically 11 rules and nodes per individual.

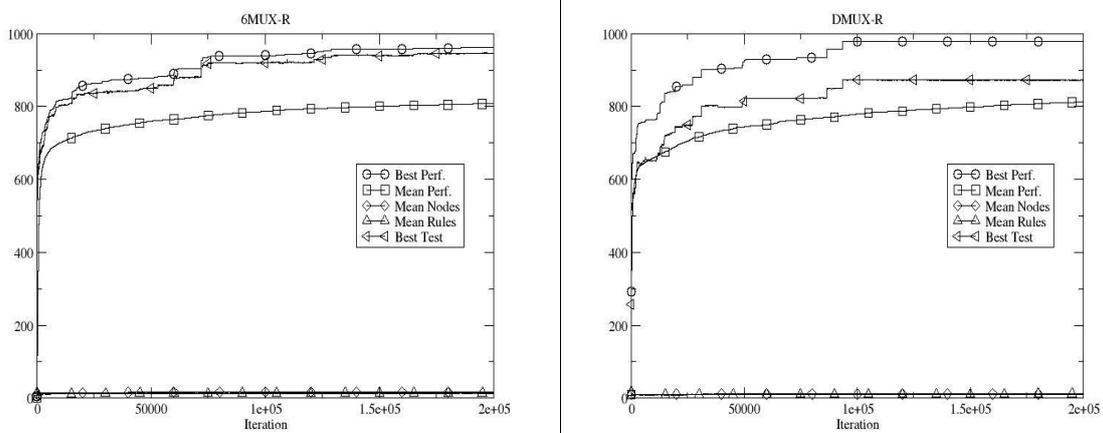

**Figure 6.** Performance on real multiplexer (left) and real demultiplexer (right).

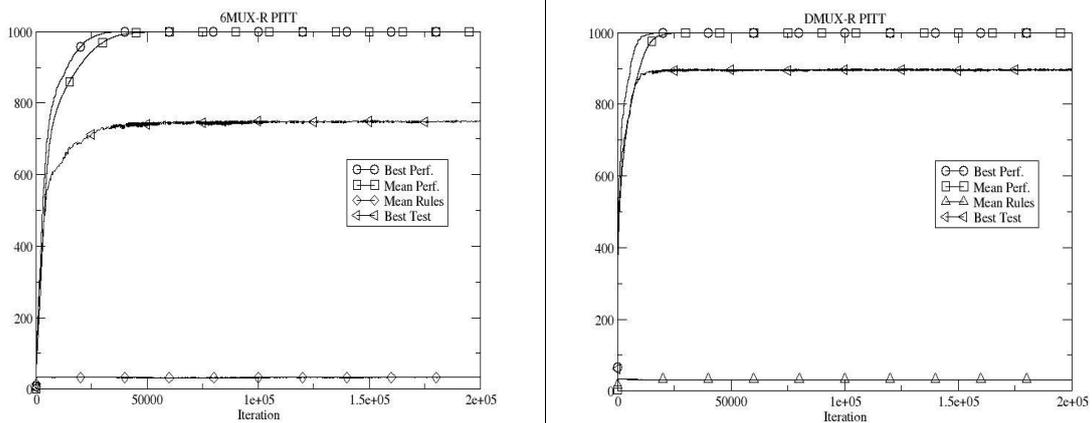

**Figure 7.** Performance of the direct encoding on real multiplexer (left) and real demultiplexer (right).

Figure 7 shows the performance of the equivalent direct encoding interval scheme on the same two tasks, with mutation operating over the same range as the imitation error, and all other details the same as before. As can be seen, performance on the test set is not significantly different for the demultiplexer (T-test, p>0.05) but significantly worse for the real multiplexer (T-test, p≤0.05). More rules were used in both cases (T-test, p≤0.05). Figure 8 shows the performance of the protein approach on the $x$=3 real multiplexer and demultiplexer. Performance with the direct encoding was significantly worse in both cases (T-test, p≤0.05), with little or no learning emerging over random behaviour for either (not shown): the benefits of the indirect encoding become clear as the task difficulty increases, not least since the search space of the traditional, direct encoding increases more rapidly with real-valued variables.

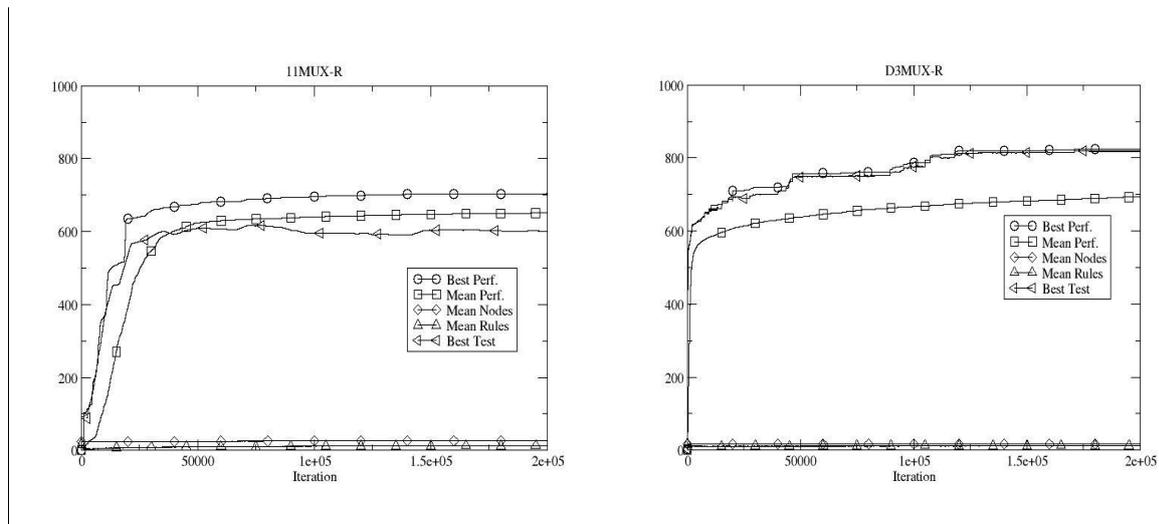

**Figure 8.** Performance on larger real multiplexer (left) and real demultiplexer (right).

## 6. Conclusions

This paper has introduced a method for learning production system rules using an indirect encoding, more specifically exploiting the dynamical behaviour of artificial genetic regulatory networks. Such general efforts exploiting temporally dynamic representations have previously been termed dynamical genetic programming [Bull, 2009]. It is suggested here that the rules may be seen as loosely analogous to the protein complexes which form within cells and process the internal and external stimuli it receives. It has been shown that the approach can be more effective than the standard way of using GRN for computation but less effective over the traditional direct encoding in binary problems. However, in continuous-valued problems the use of a restricted form of fuzzy logic within the GRN has been shown to be more effective than the direct encoding. Parameter sensitivity and other data sets are currently being explored.

Future work should consider comparison with other approaches to the evolution of rules (e.g., [Bacardit & Garrell, 2003]) and data mining algorithms, the construction of individual rules through more than one iteration of the regulatory network (e.g., one iteration = one variable for large rules), the incorporation of structural dynamism (after [Bull, 2012c]), application to multi-step problems, and allowing the protein complexes to affect the behaviour of the GRN during use (after [Bull, 2012d]), e.g., in non-Markov domains.